\documentclass[runningheads]{llncs}
\usepackage[T1]{fontenc}
\usepackage{amsmath}
\usepackage{algorithm}
\usepackage{algpseudocode}
\usepackage{subfloat}
\usepackage{subfig}
\usepackage{graphicx}
\begin{document}
\title{A Latent Space Correlation-Aware Autoencoder for Anomaly Detection in Skewed Data}
%
%
\author{Padmaksha Roy\inst{1}\orcidID{0000-0002-9571-1117} \and
Himanshu Singhal\inst{1}\orcidID{0000-0002-0474-8126} \and
Timothy J O'Shea\inst{1}\orcidID{0000-0003-2467-220X} \and
Ming Jin\inst{1}\orcidID{0000-0001-7909-4545}}
\authorrunning{P.Roy et al.}
%
\institute{Virginia Tech, VA, USA \\ 
\email{\{padmaksha,himanshusinghal,oshea,jinming\}@vt.edu}\\
}
\maketitle              
%

%
%
%
\begin{abstract}
Unsupervised learning-based anomaly detection using autoencoders has gained importance since anomalies behave differently than normal data when reconstructed from a well-regularized latent space. Existing research shows that retaining valuable properties of input data in latent space helps in the better reconstruction of unseen data. Moreover, real-world sensor data is skewed and non-Gaussian in nature rendering mean-based estimators unreliable for such cases. Reconstruction-based anomaly detection methods rely on Euclidean distance as the reconstruction error which does not consider useful correlation information in the latent space. In this work, we address some of the limitations of the Euclidean distance when used as a reconstruction error to detect anomalies (especially near anomalies) that have a similar distribution as the normal data in the feature space. We propose a latent dimension regularized autoencoder that leverages a robust form of the Mahalanobis distance (MD) to measure the latent space correlation to effectively detect near as well as far anomalies. We showcase that incorporating the correlation information in the form of robust MD in the latent space is quite helpful in separating both near and far anomalies in the reconstructed space.

\end{abstract}
\section{Introduction}
Autoencoder has an encoding network that provides a mapping from the input domain to a latent dimension and the decoder tries to reconstruct the original data from the latent dimension. Autoencoders have been successfully used in the context of unsupervised learning to effectively learn latent representations in low-dimensional space when density estimation is difficult in the high-dimensional space \cite{zong2018deep,zhou2017anomaly}. In the context of anomaly detection, it happens most of the time that the anomaly samples are rare and tedious to label making unsupervised learning the holy grail in this field. Training auto-encoders using a reconstruction error corresponds to maximizing the lower bound of mutual information between input and learned representation. Therefore, regularization methods are of great significance for preserving useful information from input space in latent space. In kernel-based autoencoders, the pairwise similarities in the data are encoded as a prior kernel matrix and the auto-encoder tries to reconstruct it from the learned latent dimension representations. This helps in learning data representations with pre-defined pairwise relationships encoded in the prior kernel matrix \cite{kampffmeyer2017deep}. Although the majority of the reconstruction-based methods assume that the outlier class cannot be effectively compressed and reconstructed, empirical results suggest that reconstruction-based approaches alone fail to capture particular anomalies that lie near the latent dimension manifold of the inlier class \cite{denouden2018improving}. Previously, authors have highlighted the challenging scenario where anomalies with high reconstruction error and residing far away from the latent dimension manifold can be easily detected but those having low reconstruction error and lying close to the manifold with inlier samples are highly unlikely to be detected. It is difficult to find a novelty score as a threshold for such cases. In this context, we try to highlight the importance of treating the far and near anomalies separately based on the feature correlation and the amount of skewness present in the data. In our case, we derive a robust form of the well-known MD distance that provides a reliable estimate of location and scale especially when the data is skewed and correlated in nature. The median and median absolute deviation(MAD) as estimators of location and scale have been studied in the robust statistics community in order to understand outliers \cite{hampel2001robust,huber2004robust}. They proved to be efficient estimators when the data follow a skewed or non-Gaussian distribution. In our latent space model, we define a strategy to deal with the near anomalies - the anomalies that lie close to the normal data in the feature space. Although autoencoders have been around for a while now, the robust autoencoder equipped with the robust MD, that we propose, can detect both near and far anomalies accurately in real-time sensor datasets which are mostly skewed and correlated in nature.

In short, our contributions can be summarized below:
\begin{itemize}
    \item We formulate a problem of detecting both near and far anomalies in real-time sensor datasets that are skewed(non-Gaussian) and have correlated features by leveraging a robust form of the well-known Mahalanobis distance that captures useful correlation information in the latent space. Our method can accurately classify those near anomalies that have a similar distribution as the normal data in the feature space while also correctly classifying the distant anomalies.
    \item Our method jointly optimizes the autoencoder reconstruction loss and the latent space regularization loss using the robust MD distance to balance the detection and generative performance of the model.
  \item Evaluation results showcase significant improvement both in terms of classification metrics and reconstruction error when experimented with different cybersecurity, and medical datasets and compared with state-of-the-art baselines.
\end{itemize}
\section{Related Work}
Unsupervised anomaly detection in latent space has been an interesting area of research for a long time. Anomaly detection with reconstruction error relies on the fact that anomalies cannot be effectively compressed and reconstructed from the latent subspace of an autoencoder when the latent space is trained on the normal data. These methods include PCA, Deep AE \cite{pang2021deep}, Robust deep AE \cite{zhou2017anomaly,yang2017joint,zhai2016deep,pang2021deep}, high dimensional anomaly detection \cite{erhan2021smart}. Widely known approaches employ a two-step process where dimensionality reduction is followed by a density estimation technique in the latent space. However, during compression, there can be a loss of useful information from the high dimensional space in the latent space. In this context, the authors \cite{zong2018deep} proposed a joint optimization of the parameters of the deep autoencoder and a mixture model for density estimation to detect anomalies in latent space. The model is trained on normal data points and then assigns low probabilities to new data points that are far from the learned normal distribution. Others propose to analyze the reconstruction error of latent dimension autoencoders and demonstrate promising results \cite{zhai2016deep}, \cite{zhou2017anomaly}. Zhou et al. proposed robust autoencoders \cite{zhou2017anomaly} that can distinguish anomalies from random noise along with discovering important non-linear features for detecting anomalies. The authors \cite{wang2019unsupervised} propose to train neural networks using unsupervised representation learning to predict data distances in a randomly projected space and then the model is optimized to learn the class structures embedded in the projected space.
However, reconstruction-based methods are limited by the fact that they do not consider any latent dimension correlation information. The authors \cite{kampffmeyer2017deep} have developed kernelized autoencoders that aim to optimize a joint objective of minimizing reconstruction error from latent space and misalignment error between prior and latent space codes which helps in detecting some kind of rare anomalies. MD-based outlier detection methods have been explored in the past for multivariate outlier detection \cite{denouden2018improving,ghorbani2019mahalanobis,ren2021simple,fort2021exploring}. Lee et al. \cite{lee2018simple} proposed an out-of-distribution (OOD) detection model using Mahalanobis distance on the features learned by a deep classification model to detect OOD samples. The authors \cite{ando2023anomaly} also emphasized the problem of overlapping of normal and anomaly class distributions in the latent space and developed a score based on how well the point fits the learned distribution of normal data. In deep probabilistic generative modeling, the anomalies or the OOD samples are detected by setting a threshold on the likelihood and selecting an efficient OOD score is often difficult. The authors of the paper \cite{xiao2020likelihood} proposed an efficient OOD score metric to detect OOD data for VAEs. The problem of near anomaly detection and OOD has been emphasized in \cite{fort2021exploring,koner2021oodformer}. In \cite{angiulli2022latent}, the authors propose detecting outliers in an augmented or enlarged latent space as anomalies tend to lie in sparse regions. In other applications \cite{zhang2018automated}, the authors propose a two-stage approach where a stacked denoising autoencoder is used to extract diverse features and followed by a KNN classifier to detect anomalies in the trained latent space. A similar kind of approach is also followed here \cite{guo2018anomaly}, where a deep neural net is used to extract the features, and a KNN classifier is leveraged in the encoded space. In addition to these, there are also other industrial applications where feature selection techniques are combined with unsupervised learning algorithms to improve anomaly detection performance \cite{rashid2022anomaly}. Our approach, on the other hand, relies on the distribution of the reconstruction error of the normal data from a regularized latent space which considers the latent dimension feature correlation in the form of robust MD distance as an additional measure to distinguish between near and far anomalies.
\footnote{\url{https://github.com/padmaksha18/DRMDIT-AE/}}
\section{Problem Formulation}
With autoencoders, appropriately modeling the distribution of the normal data is important to separate the normal and the anomaly samples in the reconstructed space. We aim to detect both near and far anomalies successfully by taking advantage of another distance metric in the latent space. Our methodology combines a robust form of the Mahalanobis Distance(MD) to measure useful feature correlation in the latent space in addition to the reconstruction loss of an AE to achieve better performance in detecting both near and far anomalies.
\subsection{ Robust Hybrid Error with MD in Latent Space} The robust MD loss using MAD and median as estimators of scale and location is derived based on the principles stated in the theorems below. We request the readers refer to the supplementary material section for detailed proof of the theorems.
\begin{theorem}
The sample median is $2/n$ times more efficient than the sample mean at exponential distribution. The result is consistent with the fact that the skewed distribution has a heavy right tail, which can cause the mean to be affected by outliers and skewness. Here, $n$ denotes the sample size.
\end{theorem}
\begin{theorem} Similarly, the relative efficiency of the sample median absolute deviation(MAD) to sample variance is $\approx$ $\frac{2}{(4 - \pi)}$ at exponential distribution. 
\end{theorem}
 The robust distance is calculated by measuring the Mahalanobis distance $\left( D_M \right)$ between the encoded samples $\left( Z \right)$ and the median of the features of the encoded samples in the latent space. The purpose of the robust MD is to effectively capture useful feature correlation information in the latent dimension feature space. The robust form of the Mahalanobis distance $\left(D_M \right)$ is calculated based on how many standard deviations an encoded sample $z_i$ is from the median of the encoded data features in the latent space. The median is calculated individually for each latent dimension feature variable. In the encoded space $\left( Z \right)$, the robust form of the MD can be estimated as
\begin{equation*}
    \hat{D}_{M} \left( Z \right) = \sqrt{\left(Z - {median}\right)^T \hat{R}^{-1} \left(Z - {median}\right)},
\end{equation*}
where $\hat{R}$ is the estimated feature-based correlation matrix of encoded data in the latent space and the robust correlation coefficient($\rho$) between two latent features $i$ and $j$ is calculated as
\begin{align*}
\rho_{{z}_{i},{z}_{j}} = \frac{\mathbf{E}[(Z_{i} - median_{i})(Z_{j} - median_{j})]}{MAD_{{z}_{i}},MAD_{{z_j}}} ,
\end{align*}
where the MAD of a latent space feature $i$ is given by
\begin{align*}
    MAD_{z_i} = median |Z_i - median(Z_i)| .
\end{align*}
Here, we aim to develop a method for estimating the feature correlation in the latent space when the data is highly skewed and features are correlated. The distance metric leverages the robust correlation estimator in the latent dimension feature space with the median and MAD as location and scale estimators respectively. On the other hand, a mean-based covariance coefficient is highly biased when the data is skewed and non-Gaussian in nature and is also unable to capture the feature correlation in the latent space without effectively estimating the standard deviation in the features.
\subsection{Objective function} The deep latent space correlation-aware autoencoder enabled with the robust MD distance is trained by minimizing the following loss function
\begin{equation}
\mathcal {L} \left(\theta,\phi \right) =  \min_{\theta,\phi} \alpha_{1}  {D}_{M}(Z_{\theta,\phi})  + \alpha_{2} 
\mathcal{L}_{e}(X, D_{\theta,\phi}(E_{\theta,\phi}(X)) ,
\end{equation}
where $D_M$ is the robust MD loss, $X$ and $Z$ are the input and latent dimensions, $\mathcal{L}_{e}$  is the reconstruction loss, $\theta$ and $\phi$ are the encoder and decoder parameters, $\alpha_1$, $\alpha_2$ are the regularization parameters that determine the weightage assigned to each part of the objective. The joint objective tries to establish a pareto optimal solution between the two objectives. We assign different weightage on robust MD loss in the latent space and the reconstruction loss and try to balance the classification and generative performance of the model.
\begin{figure*}
   \centering   \includegraphics[width=8cm,height=4cm]{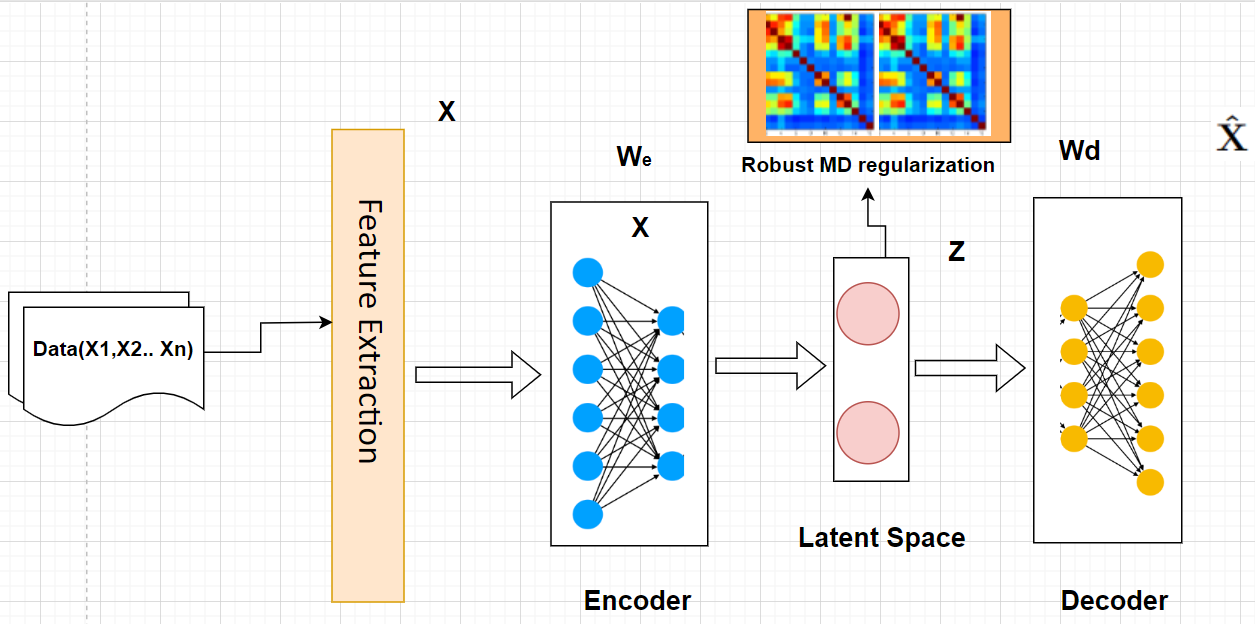}
   \caption{Deep Latent Space Correlation-Aware Autoencoder(DLSCA-AE).}
   \label{fig: DLSCA}
\end{figure*}
\section{Experiments}
Our model is a deep-stacked autoencoder model with several hidden layers and we constrain our encoder (E) and decoder (D) network to have the same architecture, that is, $W_E = W_D^{T}$. We experiment with different numbers of stacked layers for the encoder and the decoder model. During training, we use only the normal data as input to our encoder model so that the model learns a latent dimension feature representation of the normal data. The model performance is dependent on tuning two hyperparameters -- the weightage to the robust MD regularization parameter $\alpha_1$ which balances the latent dimension correlation using the robust MD and the reconstruction weightage $\alpha_2$. These two hyper-parameters must be tuned appropriately to balance the classification and generative performance of the model. 
It is difficult to detect the near anomalies using density-based methods or nearest neighbor measures as they lie very close to the normal data in the feature space. In our experiment design, we separate the anomalies into two separate classes - the near and far ones. Here, we define the near anomalies as those that are less skewed and are similar to the normal data in the feature space representation whereas the far anomalies are the ones that have highly skewed features. We refer the reader to the supplementary section to have a clear visualization of the far and near anomalies in the feature space.
\subsection{Datasets}
\begin{itemize}
 \item \textbf{CSE-CIC-IDS2018} This is a publicly available cybersecurity dataset that is made available by the Canadian Cybersecurity Institute (CIC).  We consider the data from two different days which consists of different kinds of attacks. The anomaly data is labeled as '1'. 
 \item \textbf{NSL-KDD} This is also a publicly available benchmark cybersecurity dataset made available by CIC. It has a total of 43 different features of internet traffic flow. We use 21 correlated features to develop our model. 
\item \textbf{Arrythmia} It is a multi-class classification dataset and the aim is to distinguish between the presence and absence of cardiac arrhythmia and to classify it in one of the normal or anomaly groups.
\end{itemize}
\begin{table}[]
\centering
\setlength{\tabcolsep}{1.5pt}
\small
    
\label{tab:results}
        \begin{tabular}{c|c|c|c|c}
        Datasets & Dimensions & Corr Feats & Samples(Train) & Anomaly ratio(Test)
        \\ \hline
        CSE-CIC-IDS & 79 & 29 & 50000 & 0.5 
        \\ 
        NSL-KDD & 43 & 20 & 50000 & 0.5 
        \\
        Arrythmia & 34 & 20 & 50000 & 0.5
        \\ 
        \hline
        \end{tabular}
        \normalsize
\end{table}
\begin{figure}[htp!]%
  \subfloat[\centering]{{\includegraphics[height=3cm,width=3cm]{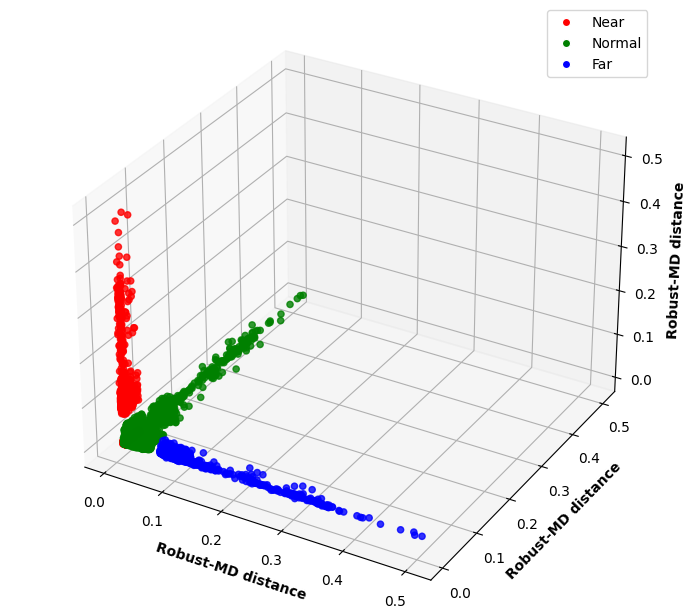}}}%
    \subfloat[\centering]{{\includegraphics[height=3cm,width=3cm]{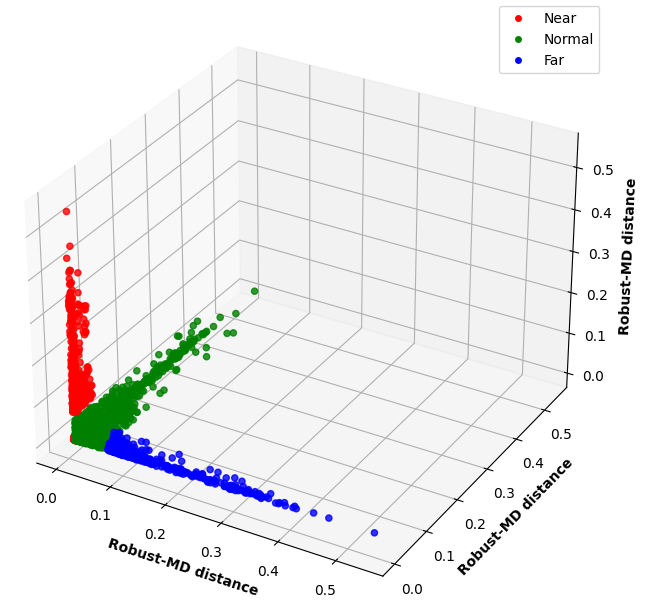}}}%
   \subfloat[\centering]{{\includegraphics[height=3cm,width=3cm]{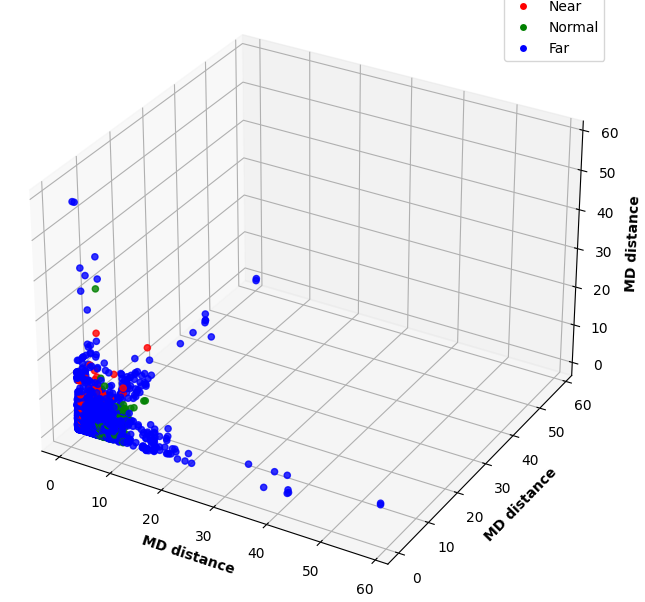}}}%
   \subfloat[\centering]{{\includegraphics[height=3cm,width=3cm]{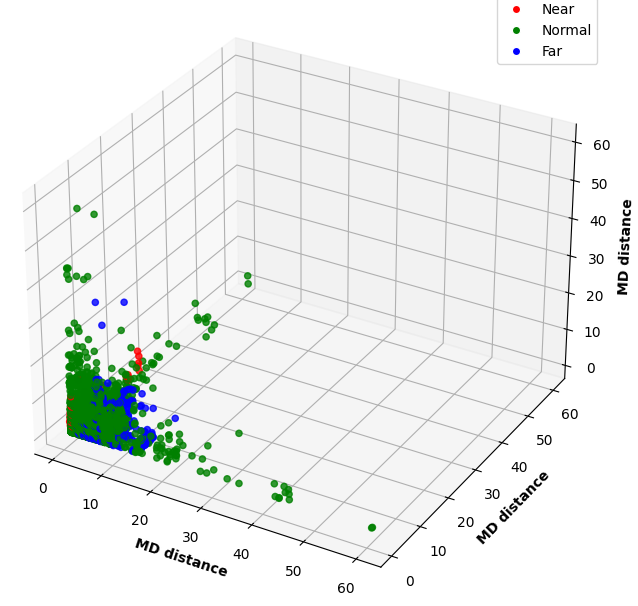}}}%
    \caption{(a),(b) shows the reconstructed space when the robust MD is used as a regularizer and (c),(d) corresponds to the reconstructed space with standard MD with mean and covariance as regularizer. }
   \label{Anomaly_detect_recons_space}
   \end{figure}
\subsection{Baseline Methods}
In order to compare the performance of our model, we consider standard baselines like kernel-based autoencoders(DKAE), density estimation models(DAGMM), VAE and, AE, AE-KNN, mean-based MD Autoencoder, and some statistical models.
\begin{itemize}
 \item \textbf{DKAE} \cite{kampffmeyer2017deep} The Deep Kernelized Autoencoder \cite{kampffmeyer2017deep} has a kernel alignment loss that is calculated as the normalized Frobenius distance between the latent dimension code matrix and the prior kernel matrix and a reconstruction loss.
 \item \textbf{DAGMM}\cite{zong2018deep} The Deep Autoencoding Gaussian Mixture Model \cite{zong2018deep} is an unsupervised anomaly detection model that optimizes the parameters of the deep autoencoder and the mixture model simultaneously using an estimation network to facilitate the learning of a Gaussian Mixture Model (GMM). 
  \item \textbf{VAE} VAE leverages a probabilistic encoder-decoder network and the reconstruction probability is used for detecting anomalies. It also tries to regularise the organization of the latent space by making the distributions returned by the encoder close to a standard Gaussian distribution. 
  \item \textbf{MD-based Autoencoder} \cite{denouden2018improving}This model also leverages MD in the latent space with mean as the estimator of location and the sample covariance. 
  \item \textbf{PCA} PCA performs a linear transformation to convert a set of correlated variables into a set of linearly unrelated correlated variables of a smaller dimension. Its primary aim is to perform dimensionality reduction and it achieves it by identifying the principal components along which the data points vary the most.
  \item \textbf{OCSVM} One class SVM is a popular anomaly detection algorithm that finds the decision boundary of the normal data based on SVM with kernel approximation and separates the data from the origin in the transformed high-dimensional predicted space. 
  \item \textbf{Stacked AE} Autoencoders are encoder-decoder neural networks that learn a compressed latent space representation of the normal data. The representation of the anomaly data deviates significantly from the learned latent space and thus has a higher reconstruction error.
  \item \textbf{Isolation Forest} Isolation Forest builds a random ensemble of trees by randomly selecting features and splitting them based on a chosen split value. Anomalies are expected to have a smaller depth, i.e, distance between the root and the nodes as they have lesser and distinct and can be separated easily.
  \item \textbf{Stacked AE-KNN} \cite{guo2018anomaly} Autoencoders are used to perform dimensionality reduction upon the data and then the K-nearest neighbors algorithm is applied on the latent space representation obtained from the autoencoder. 
\end{itemize}
\begin{table*} 
\begin{tabular}{lcccccccccc}
\hline {} & {} &  \multicolumn{4}{c}{ CSE-CIC-IDS-1 Dataset } & \multicolumn{4}{c}{ NSL-KDD}
\\
\cline{3-6} 
\cline{7-10}

 Model & Type  & Accuracy & Precision & Recall & AUC  & Accuracy & Precision  & Recall &AUC \\

\hline DLSCA-AE & Near & $\mathbf{91.5}$ & $\mathbf{90.7}$ & ${93.4}$ & $\mathbf{94.5}$&

${92.4}$&
$\mathbf{94.6}$&
$\mathbf{93.1}$&
$\mathbf{96.0}$&
\\

& 
Far & ${98.9}$ & ${90.1}$ & ${92.4}$ & ${98.9}$

 & ${91.2}$& ${95.2}$ &

$95.4$ & ${97.3}$ & 
\\

\hline DKAE & Near & $50$ & ${75.2}$ & $50.8$ & $71.4$ &

$82.0$ & $81.6 $ &
$92.2$ & $95.6$  \\

 & Far & $79.8$ & ${85.2}$ & $79.8$ &$74.5$

& $74.5$ & $74.7$ & $74.7$ & $74.5$  \\

\hline

\hline DAGMM & Near & $67.6$ & $68.1$ & $67.9$ & 
$67.9$

 & $\mathbf{95.4}$ & $86.9$ & $89.3$ & $88.8$ \\

 & Far & $66.6$ & $75.6$ & $66.8$ & $65.1$

 & ${94}$ & $86.8$ & $89.1$ & $91.9$ \\





\hline MD-AE & Near & $57.3$ & $57.2$ & $52.2$ & $53.4$

 & $52.3$ & $53.9 $ &
$52.5$ & $62.6$ \\

 & Far & $52.8$ & $50.6$ & $51.6$ & 71.6

 & $51.3$ & $53.6$ & 

$53.5$ & $69.5$ \\

\hline PCA 
 & Near & $52.2$ & $51.1$ & $96.2$ & $18.6$

 & $78.5$ & $73.7$ & 

$88.4$ & $72.8$ \\

& Far & $99.8$ & $99.6$ & $99.9$ & $99.9$

 & $93.9$ & $99.5$ &
$88.2$ & $93.8$ \\

\hline OCSVM 
& Near & $63.7$ & $83.7$ & $34.0$ & $60.1$

 & $97.1$ & $97.1$ & 

$97.0$ & $99.5$ \\

& Far & $97.7$ & $\mathbf{98.9}$ & $96.4$ & $99.6$

 & $98.4$ & $99.7$ &
$96.8$ & $99.7$ \\

\hline Stacked AE 
 & Near & $50.0$ & $0.0$ & $0.0$ & $2.9$

 & $83.1$ & $82.3$ & 

$84.2$ & $87.4$ \\

& Far & $93.9$ & $97.0$ & $90.6$ & $98.3$

 & $96.8$ & $94.8$ &
$99.0$ & $99.5$ \\

\hline VAE 
 & Near & $49.9$ & $0.0$ & $0.0$ & $1.9$

 & $82.8$ & $81.2$ & 

$85.2$ & $85.6$ \\

& Far & $\textbf{99.9}$ & $99.8$ & $99.9$ & $99.9$

 & $95.7$ & $92.3$ &
$99.6$ & $99.1$ \\

\hline Isolation Forest 
 & Near & $84.8$ & $79.4$ & $\mathbf{93.8}$ & $80.5$
 & $90.1$ & $86.7$ & 
$94.6$ & $94.9$ \\

& Far & $94.0$ & $96.2$ & $91.6$ & $97.6$
 & $95.8$ & $93.5$ &
$98.4$ & $98.6$ \\






\hline Stacked AE-KNN 
& Near & $82.7$ & $90.7$ & $72.8$ & $82.4$
 & $84.7$ & $83.3$ & 
$86.2$ & $88.0$ \\
& Far & $97.4$ & $96.1$ & $\textbf{98.8}$ & $\textbf{99.1}$
 & $\textbf{98.3}$ & $\textbf{98.5}$ &
$\textbf{98.0}$ & $\textbf{99.8}$ \\

\hline
\end{tabular}

{
\begin{tabular}{lcccccccccc}
\hline {} & {} &  \multicolumn{4}{c}{ CSE-CIC-IDS-2 } & \multicolumn{4}{c}{ Arrythmia}
\\
\cline{3-6} 
\cline{7-10}

 Model & Type  & Accuracy & Precision & Recall & AUC  & Accuracy & Precision  & Recall &AUC \\
\hline DLSCA-AE & Near & $\mathbf{91.4}$ & ${91.7}$ & ${93.4}$ & $\mathbf{94.5}$&

$\mathbf{92.4}$&
${94.6}$&
$\mathbf{93.1}$&
$\mathbf{96.0}$&
\\
& 
Far & ${98.9}$ & ${90.1}$ & ${92.4}$ & ${98.9}$
 & ${91.2}$& $95.2$ & 
${95.4}$ & ${97.3}$ & 
\\
\hline DKAE & Near & $50$ & $ {75.2}$ & $50.8$ & $71.4$ &$82.0$ & $81.6 $ &
$92.2$ & $95.6$  \\
& Far & $79.8$ & ${85.2}$ & $79.8$ &$74.5$ 
& $74.5$ & $74.7$ & $74.7$ & $74.5$  \\
\hline
\hline DAGMM & Near & $67.6$ & $68.1$ & $67.9$ & 
$67.9$
& $81.64$ & $\mathbf{94.72}$ & $66.46$ & $78.11$ \\
& Far & $66.6$ & $75.6$ & $66.8$ & $65.1$
& $99.1$ & $99.2$ & $99.2$ & $99.2$ \\
\hline MD-AE & Near & $57.3$ & $57.2$ & $52.2$ & $53.4$
& $52.3$ & $53.9 $ &
$52.5$ & $62.6$ \\
& Far & $52.8$ & $50.6$ & $51.6$ & 71.6
& $51.3$ & $53.6$ & 
$53.5$ & $69.5$ \\
\hline PCA 
 & Near & $51.8$ & $50.9$ & $96.5$ & $18.2$
 & $87.3$ & $95.9$ & 
$77.5$ & $83.0$ \\
& Far & $99.8$ & $99.6$ & $99.9$ & $99.9$
 & $99.7$ & $99.8$ &
$\mathbf{99.9}$ & $99.8$ \\
\hline OCSVM 
& Near & $56.5$ & $80.2$ & $17.2$ & $56.6$
& $81.6$ & $90.4$ & 
$70.1$ & $80.8$ \\
& Far & $98.4$ & $\mathbf{99.0}$ & $97.9$ & $99.7$
& $99.6$ & $99.7$ &
$96.8$ & $\mathbf{99.7}$ \\
\hline Stacked AE 
 & Near & $49.9$ & $0.0$ & $0.0$ & $3.3$
& $66.7$ & $93.8$ & 
$34.7$ & $50.9$ \\
& Far & $66.6$ & $66.6$ & $99.9$ & $74.2$
& $99.2$ & $99.3$ &
$99.2$ & $99.4$ \\
\hline VAE 
 & Near & $49.9$ & $0.0$ & $0.0$ & $2.2$
 & $71.9$ & $90.9$ & 
$47.7$ & $64.7$ \\
& Far & $\mathbf{99.7}$ & $99.5$ & $\mathbf{99.9}$ & $\mathbf{99.9}$
 & $\mathbf{99.7}$ & $\mathbf{99.8}$ &
$99.6$ & $99.7$ \\
\hline Isolation Forest 
 & Near & $84.9$ & $79.1$ & $\mathbf{94.9}$ & $79.9$
 & $91.5$ & $90.3$ & 
$92.7$ & $97.3$ \\
& Far & $93.7$ & $97.5$ & $89.6$ & $97.4$
 & $99.9$ & $99.8$ &
$99.9$ & $99.9$ \\
\hline Stacked AE-KNN 
& Near & $80.5$ & $\mathbf{92.8}$ & $66.1$ & $77.2$
 & ${88.2}$ & $88.1$ & 
$77.3$ & $99.7$ \\
& Far & $96.0$ & $95.9$ & $96.0$ & $98.8$
 & $83.7$ & $88.5$ &
$77.4$ & $79.8$ \\
\hline
\end{tabular} }
  \caption{Here, we compare the performance of the proposed model(DLSCA-AE) and baseline models while detecting near and far anomalies.}
\end{table*}

\subsection{Ablation Study} 
 During training, we use only the normal data as input to our encoder model so that the model learns a latent dimension feature representation of the normal data. We consider the standard classification metrics such as accuracy, precision, and recall to demonstrate the detection performance of the model. We kept the training distribution comprising of 50k normal data samples that are used as training data and the test data consists of an equal proportion of normal and anomaly OOD samples. The near anomalies while getting reconstructed from the latent space which is regularized with the robust MD distance of the normal data are projected to a different subspace in the reconstructed space which efficiently helps them to be separated from the normal data in the reconstructed space. The reconstructed space as depicted in Fig 2, shows that the robust MD regularized AE can separate both near and far anomalies from the normal data compared to an MD regularized AE with mean and sample covariance estimator. The threshold for anomaly detection depends on the distribution of the reconstruction distance of the normal data. We choose the threshold for anomaly detection to be two standard deviations from the mean of the reconstructed distance of the normal data. In Fig \ref{Anomaly_detect_recons_space}, we try to detect the near and far anomalies using a threshold that is chosen based on the reconstruction range of the normal data.  Our proposed autoencoder DLSCA-AE showcases an improvement of ${5\%-15\%}$ in MSE and ${5\%-8\%}$ in MAE  while reconstructing unseen data compared to the DKAE model which uses simple Euclidean distance as reconstruction error. Better MSE and MAE of reconstruction do not just imply good learning of the input representations in the latent space but also the performance of the encoder in back-mapping to the higher dimensional space. Due to the good generative performance of the model, it can be also leveraged to generate synthetic data in applications where data is not easily available. 
\begin{table*}
\centering
{
\begin{tabular}{lcccccccccc}
\hline {} & {} &  \multicolumn{2}{c}{ CSE-CIC-IDS } & \multicolumn{2}{c}{ NSL-KDD}
\\
\cline{3-6} 
\cline{7-10}

 Model & Type  & MSE & MAE & MSE & MAE \\

\hline DLSCA-AE & Near & $\mathbf{0.173 \pm 0.008}$ & $\mathbf{0.229 \pm 0.006}$ & $\mathbf{0.935 \pm 0.003}$ & $\mathbf{0.722 \pm 0.003}$&

\\
& 
Far &${1.413 \pm 0.020}$ & ${0.760 \pm 0.004}$ & $\mathbf{0.935 \pm 0.040}$ & ${0.890 \pm 0.009}$

\\

\hline DKAE & Near & $0.334 \pm 0.005$ & $0.396 \pm 0.005$ & $1.686 \pm 0.008$ & $0.890 \pm 0.002 $ &
 \\

 & Far & $1.870 \pm 0.020$ & $0.950 \pm 0.030$ 
& $1.673 \pm 0.050$ & $0.913 \pm 0.005$  \\

\hline DAGMM & Near & $0.258 \pm 0.006$ & $0.390 \pm 0.003$ & $ 1.790 \pm 0.005$ & $ 0.998 \pm 0.005$ &
 \\

 & Far & $ 1.454\pm 0.030$ & $0.7925 \pm 0.010$ 
& $1.6006 \pm 0.040$ & $0.903 \pm 0.040$  \\

\hline AE & Near & $0.183 \pm 0.006$ & $0.304 \pm 0.003$ & $ 1.302 \pm 0.009$ & $0.815 \pm 0.005 $ &
 \\

 & Far & $0.775 \pm 0.030$ & $0.562 \pm 0.010$ 
& $1.083 \pm 0.040$ & $\mathbf{0.743} \pm 0.008$  \\

\hline VAE & Near & $0.202 \pm 0.009$ & $0.383 \pm 0.003$ & $1.280 \pm 0.006$ & $0.832 \pm 0.003 $ &
 \\
 & Far & 
 $\mathbf{1.411 \pm 0.020}$ & $\mathbf{0.673 \pm 0.030}$ 
& $1.051 \pm 0.050$ & $0.775 \pm 0.005$  \\
\hline
\end{tabular} }
 \caption{It shows the comparison of reconstruction MSE and MAE with the baselines}
\end{table*}
\section{Hyperparameter Sensitivity}
While training, we put more weightage on the MD-based regularization($\alpha_1$) and less weightage on the reconstruction error($\alpha_2$). We train our model up to 150-200 epochs. Batch size is another important training hyper-parameter that directly relates to the number of samples used to estimate the correlation matrix in the latent space. We observe that the initial training is more stable when we use a higher number of samples to estimate the feature correlation in the latent space. After experimenting with different weightage for both the reconstruction loss ($\alpha_2$) and the MD regularizer ($\alpha_1$), we find that the best results in terms of classification metrics are obtained when $\alpha_2$ are in the range of \{0.1 to 0.2 \} and $\alpha_1$ is chosen between \{0.8 to 0.9\}. Otherwise, we suggest keeping the reconstruction weightage significantly less than the robust MD regularizer to determine a good threshold for detecting both near and far anomalies. The best results are reported in Table 1. We showcase the effectiveness of the latent space regularization in improving the generative performance of the model with both near and far anomalies in Table 2.
\begin{figure}[htp!]%
  \subfloat[\centering]{{\includegraphics[height=3cm,width=3cm]{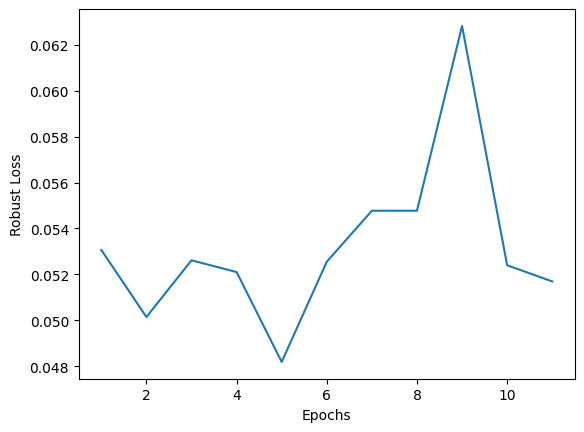}}}%
    \subfloat[\centering]{{\includegraphics[height=3cm,width=3cm]{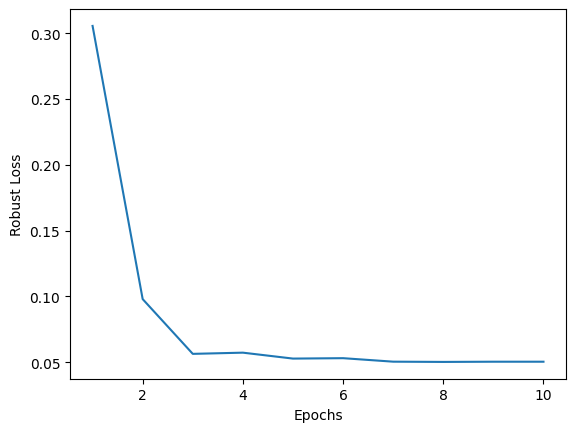}}}%
   \subfloat[\centering]{{\includegraphics[height=3cm,width=3cm]{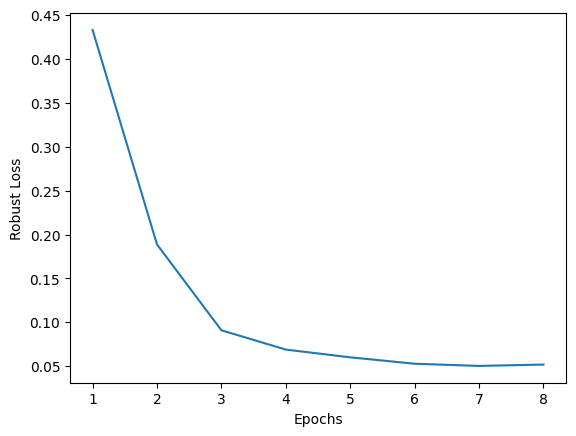}}}%
   \subfloat[\centering]{{\includegraphics[height=3cm,width=3cm]{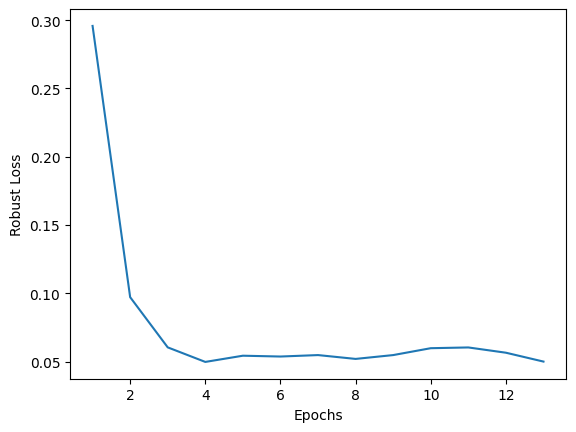}}}%
    \caption{(a), (b), (c), (d) shows the validation error during training with sample sizes 50, 100, 200, and 300 respectively. We see that the training is more stable when the batch size is higher during each epoch of training.  }
   \label{Batch Size Sensitivity}
   \end{figure}
\section{Conclusion} In this paper, we propose a correlation-aware regularized autoencoder that leverages a robust form of MD to capture correlation information in the latent space which helps to detect certain kinds of anomalies more efficiently. The MAD- and median-based robust correlation estimators are useful indicators of specific kinds of anomalies especially when the data is skewed and features are correlated. We find significant improvement in detecting the near anomalies while having equally good performance in detecting the far anomalies with standard classification metrics on standard datasets. We also showcase the improvement of the generative performance of the model while reconstructing data. In the future, we would be interested in understanding the effectiveness of bounding the robust correlation between related data domains to achieve out-of-distribution(OOD) generalization.
\section*{Acknowledgment} We would like to thank Virginia Tech National Security Institute(VTNSI) for supporting our work and Dr. Lamine Mili (ECE, Virginia Tech) for the introductory course on Robust Statistics.

\title{Supplementary Material\thanks{Supported by organization x.}}
\section{Proof of Theorem 1}
\begin{proof}
Let us derive the variance of the sample median for exponential(right-skewed) distribution. For an exponential distribution with parameter $\lambda$, the probability density function is given by:
\begin{equation*}
    f(x) = \lambda \exp^{-\lambda x}, \hspace{2mm}\text{for} \hspace{2mm} x \ge 0
\end{equation*}
The cumulative distribution function is given by,
\begin{equation*}
    F(x) = 1 - \exp^{-\lambda x}, \hspace{2mm} \text{for} \hspace{2mm} x \ge 0 
\end{equation*}
The median of the distribution is given by
\begin{equation*}
    m = \frac{\ln(2)}{\lambda}
\end{equation*}
To find the variance of the sample median, we can use the formula
\label{var_med}
\begin{equation}
    Var(M) \approx \frac{\pi}{2} \frac{1}{F(m)^2}Var(X) ,
\end{equation}
where, M is the sample median, $X$ is a random variable from the population distribution, $F(m)$ is the cumulative distribution function of $X$ evaluated at the median value m.
In order to calculate F(m),
\begin{equation*}
\begin{split}
    F(m) = 1 - \exp^{-\lambda m} \\
    &= 1 - \exp^{(-\lambda \ln(2)/\lambda)} \\ &= 1 - 1/2 \\ &= 1/2 
    \end{split}
\end{equation*}
Now, we calculate the variance of the random variable X  which is equal to the square of the population standard deviation, given by,
\begin{equation}
    Var(X) = \sigma^2 = \frac{1}{\lambda^2}
\end{equation}
Finally, if we substitute the values in \eqref{var_med},

\begin{equation}
\begin{split}
    Var(M) \approx \frac{\pi}{2} \frac{1}{F(m)^2} Var(X) \\
    &= \pi/2 \cdot 2^2 \cdot \frac{1}{\lambda^2} \\&= \frac{1}{2 \lambda^2}
\end{split} 
\end{equation}
Therefore, the variance of the sample median for an exponential distribution is approximately equal to $\frac{1}{2 \lambda^2}$. This means that the variance of the sample median decreases as the rate parameter $\lambda$ increases. 
Now, let us derive the variance of the sample mean for exponential distribution. For an exponential distribution with parameter $\lambda$, the probability density function is given by:
\begin{equation}
    f(x) = \lambda \exp^{-\lambda x}, \hspace{2mm}\text{for} \hspace{2mm} x \ge 0
\end{equation}
The cumulative distribution function is given by,
\begin{equation*}
F(X) = 1 - \exp^{-\lambda x}, \hspace{2mm}\text{for} \hspace{2mm} x \ge 0
\end{equation*}
The mean of the distribution is given by,
\begin{equation*}
    \mu = \frac{1}{\lambda}
\end{equation*}
In order to find the variance of the sample mean, 
\begin{equation*}
    Var(\bar{X}) = \frac{Var(X)}{n},
\end{equation*}
where, $\bar{X}$ is the sample mean, X is a random variable from the population distribution, Var(X) is the variance of X, and n is the sample size.
The variance of X is given by,
\begin{equation*}
    Var(\bar{X}) = \frac{Var(X)}{n},
\end{equation*}
where, $\bar{X}$ is the sample mean, X is a random variable from a population distribution, Var(X) is the variance of X and n is the sample size.
The variance of X is given by,
\begin{equation*}
    Var(X) = E[X^2] - (E[X])^2
\end{equation*}
The expected value of X is given by,
\begin{equation*}
    E[X] = \frac{1}{\lambda}
\end{equation*}
The expected value of $X^2$ is,
\begin{equation*}
\begin{split}
    E[X^2] =  \int_{0}^{\infty} x^2 \cdot \lambda \cdot \exp^{-\lambda x}\,dx 
    &= \frac{2}{\lambda^2}
    \end{split}
\end{equation*}
Therefore, the variance of X is:
\begin{equation*}
\begin{split}
Var(X) = E[X^2] - (E[X])^2 \\
&= \frac{2}{\lambda^2} - \frac{1}{\lambda^2} \\
&= \frac{1}{\lambda^2}
 \end{split}   
\end{equation*}
Substituting this into the formula of the variance of the sample mean, we get, 
\begin{equation*}
\begin{split}
    Var(\bar{X}) = \frac{Var(X)}{n}
    &= \frac{1}{n\lambda^2}
\end{split}
\end{equation*}
Therefore, the variance of the sample mean for an exponential distribution is equal to,
\begin{equation*}
    Var(\bar{X}) = \frac{1}{n\lambda^2}
\end{equation*}
\begin{equation*}
    Var(\bar{M}) = \frac{1}{2 \lambda^2}
\end{equation*}
Therefore, the relative efficiency of the sample median to the sample mean is,
\begin{equation*}
    \frac{Var(\bar {X})}{Var(\bar{M})} = \frac{\frac{1}{n \lambda^2}}{\frac{1}{2 \lambda^2}} 
    = \frac{2}{n}
\end{equation*}

This means that, for an exponential distribution which is a skewed form of distribution, the sample median is on average 2/n times more efficient than the sample mean in terms of variance. In other words, if we use the sample median instead of the sample mean, we can achieve the same level of precision with a sample size that is only half as large. This result is consistent with the fact that the exponential distribution has a heavy right tail, which can cause the mean to be affected by outliers and skewness. The median, on the other hand, is a robust measure of central tendency that is less affected by extreme values.
\end{proof}
\section{Proof of Theorem 2}
\begin{proof}Let $X$ be a random variable with an exponential distribution with mean $\mu$. Then the variance of $X$ is given by,
\begin{equation*}
    Var(X) = \mu^2
\end{equation*}
Now, let MAD be the median absolute of $X$. The formula for MAD  is given by,
\begin{equation*}
    MAD = median(|x - \mu|)
\end{equation*}
where, median denotes the median value of the set of absolute deviations.
Since $X$ has an exponential distribution, we can write the CDF of X as:
\begin{equation*}
    F(X) = 1 - \exp^{-\lambda x}
\end{equation*}
where, $\lambda = \frac{1}{\mu}$ is the rate parameter of the exponential distribution. 
Now, the CDF of $|x - \mu|$ is given as,
\begin{equation}
\begin{split}
    F(|X -\mu|) = P(|X - \mu| \leq x)\\
    &=F(X + \mu) - F(\mu -X)\\
    &= 2 F(x) - 1
\end{split}
\end{equation}
where, $0 \leq x \leq \mu$
The median absolute deviation MAD is the median of the set of absolute deviations $|x - \mu|$. Therefore, we need to find the value of $x$ such that, $F(|x - \mu| = 1/2)$. Solving for $x$, we get,
\begin{equation*}
\begin{split}
    |x - \mu| = F^{-1}(1/2) \\
    &= \frac{\ln(2)}{\lambda}
\end{split}
\end{equation*}
Therefore, the median absolute deviation is,
\begin{equation*}
    MAD = |x - \mu| = \frac{\ln 2}{\lambda}
\end{equation*}
To compute the variance of MAD, we can write the formula
\begin{equation*}
    Var(MAD) = \mathbf{E}((|x - \mu| - MAD)^2),
\end{equation*}
Now, expanding the square, we get, 
\begin{equation*}
    Var\left(MAD\right) = \mathbf{E}\left(|x - \mu|^2\right) - 2 MAD \{\mathbf{E} \left(|x - \mu|\right)\} + {MAD}^2
\end{equation*}
Since, $X$ has exponential distribution, the expectations can be computed as,
\begin{equation*}
    \mathbf{E}(|x - \mu|^2 = 2 \mu ^2
\end{equation*}
\begin{equation*}
    \mathbf{E}(|x - \mu|= \frac{2}{\lambda}
\end{equation*}
\begin{equation*}
    MAD^2 = \left(\frac{\ln(2)}{\lambda}\right)^2
\end{equation*}
Substituting these expressions, we get, 
\begin{equation*}
    Var(MAD) = 2 \mu^2 - 4 \mu \frac{\ln(2)}{\lambda} + \frac{\ln^2(2)}{\lambda^2}
\end{equation*}
Now, the relative efficiency of MAD over variance for the exponential distribution is given by,
\begin{equation*}
\begin{split}
 \frac{Var(X)}{Var(MAD)} = \frac{\mu^2}{ \mu^2 - 4 \mu \frac{\ln(2)}{\lambda} + \frac{\ln^2(2)}{\lambda^2} } 
 &= \frac{2}{4 - \pi}
\end{split}
\end{equation*}

\end{proof}

%
%
%
%





\section{Histograms of Features and Hyper-parameter sensitivity.}
\begin{figure}  \includegraphics[width=16cm,height=12cm]{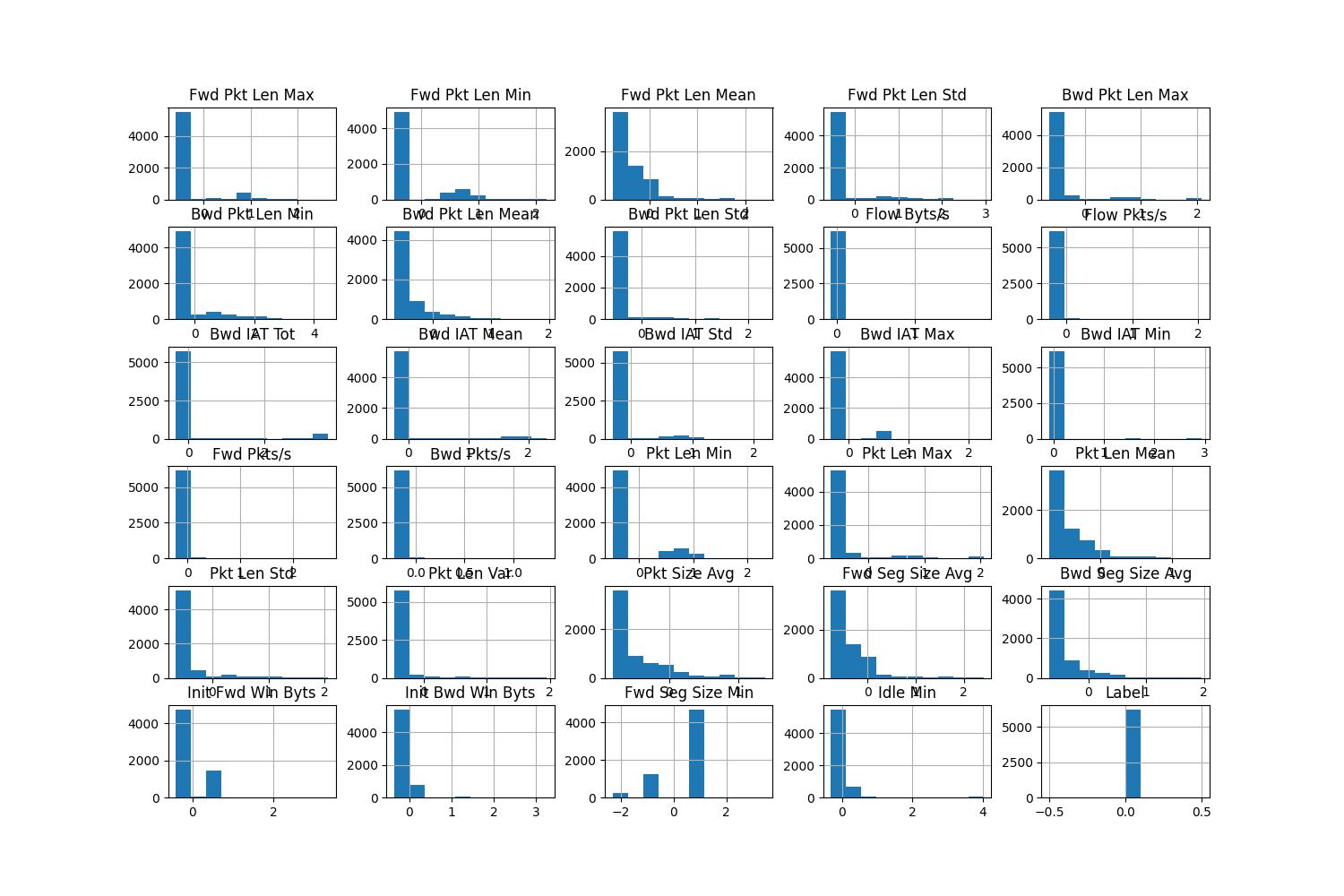}
  \caption{Histogram of normal data features of CSE-CIC-IDS dataset.}
\end{figure}
\begin{figure}
  \centering  \includegraphics[width=16cm,height=12cm]{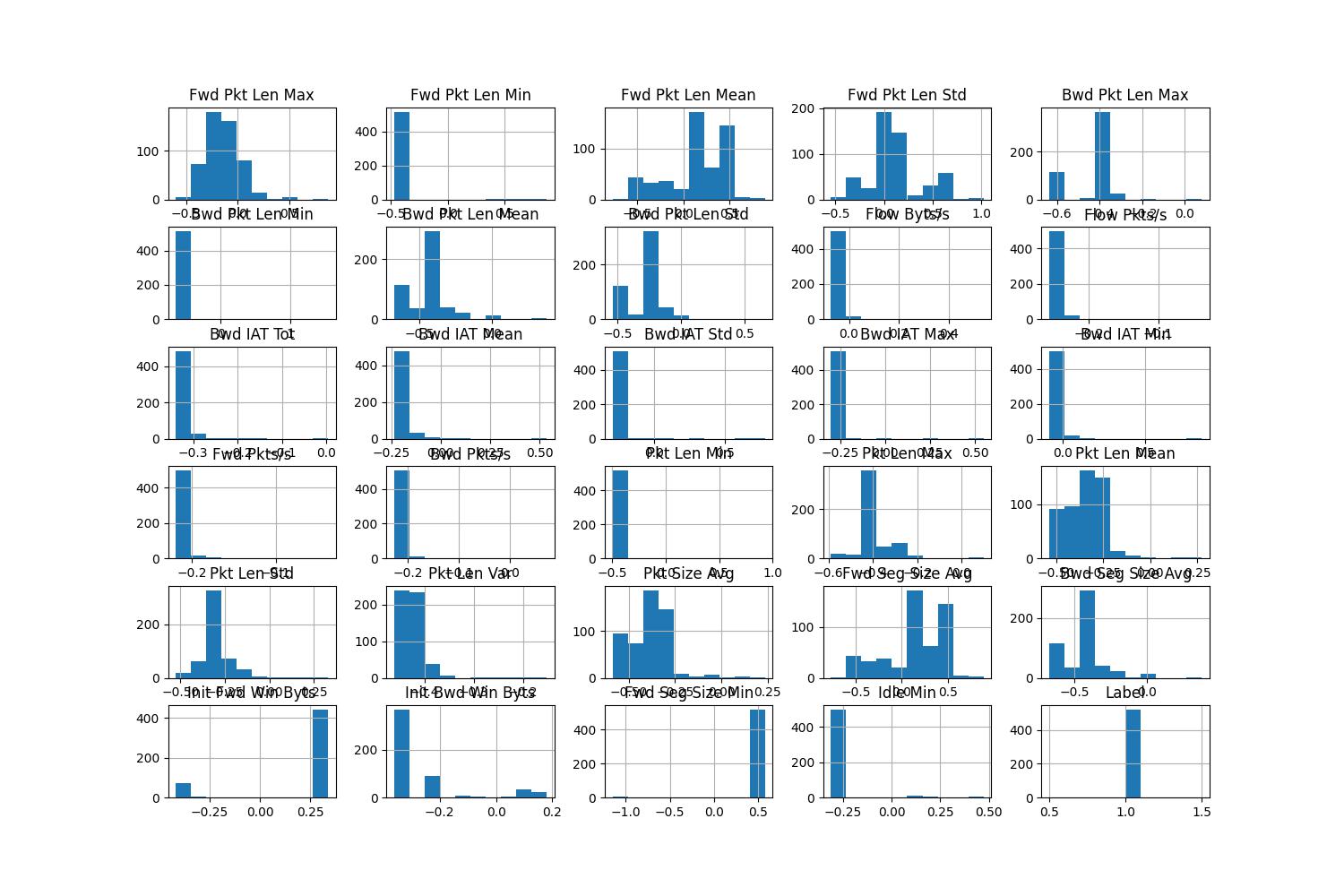}
  \caption{Histogram of near anomaly features(less skewed) in CIC-CSE-IDS dataset.}
\end{figure}
\begin{figure}
  \centering
  \includegraphics[width=16cm,height=12cm]{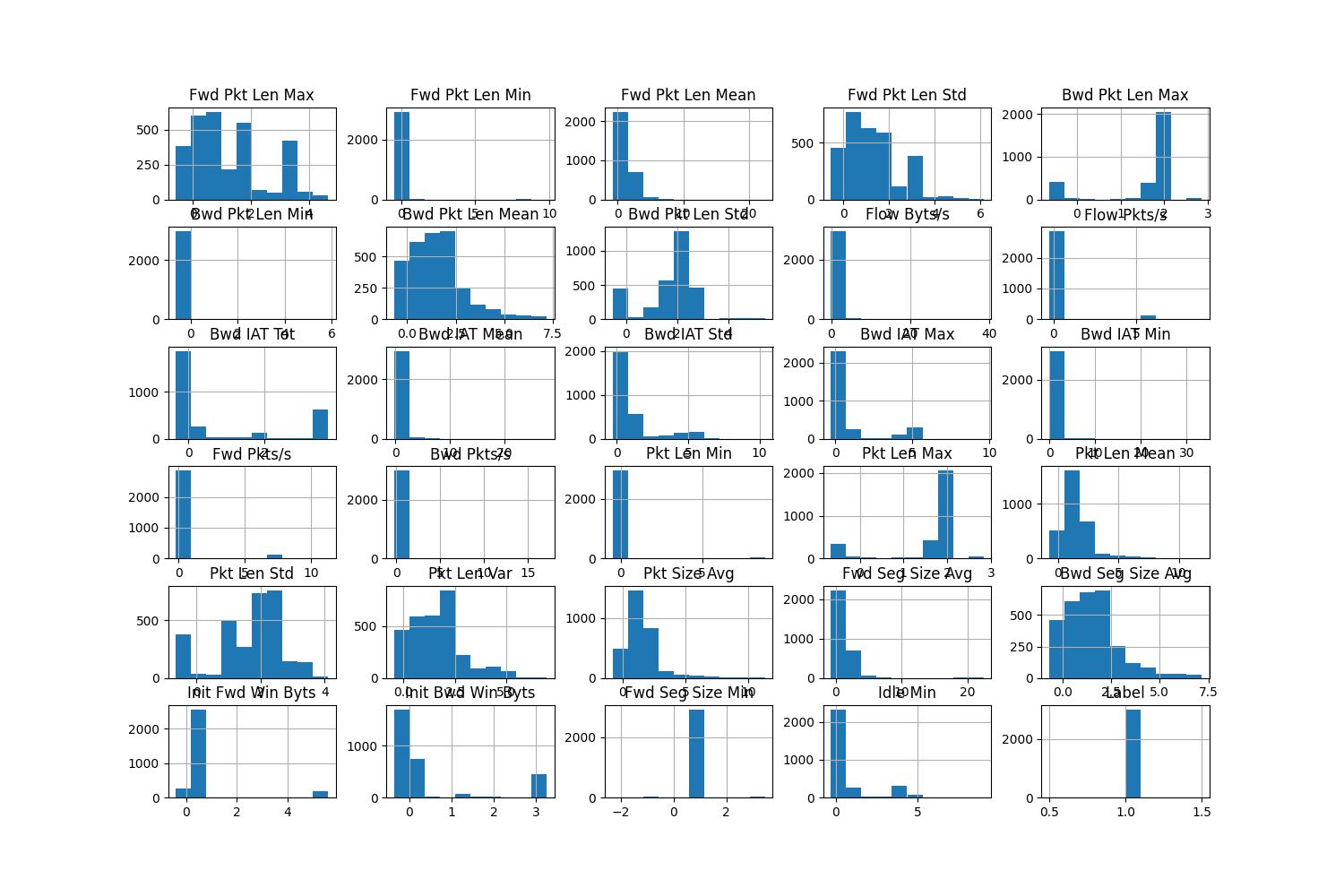}
  \caption{Histogram of far anomaly features(highly-skewed) in CIC-CSE-IDS dataset.}
\end{figure}
\begin{figure}
  \centering
  \includegraphics[width=16cm,height=12cm]{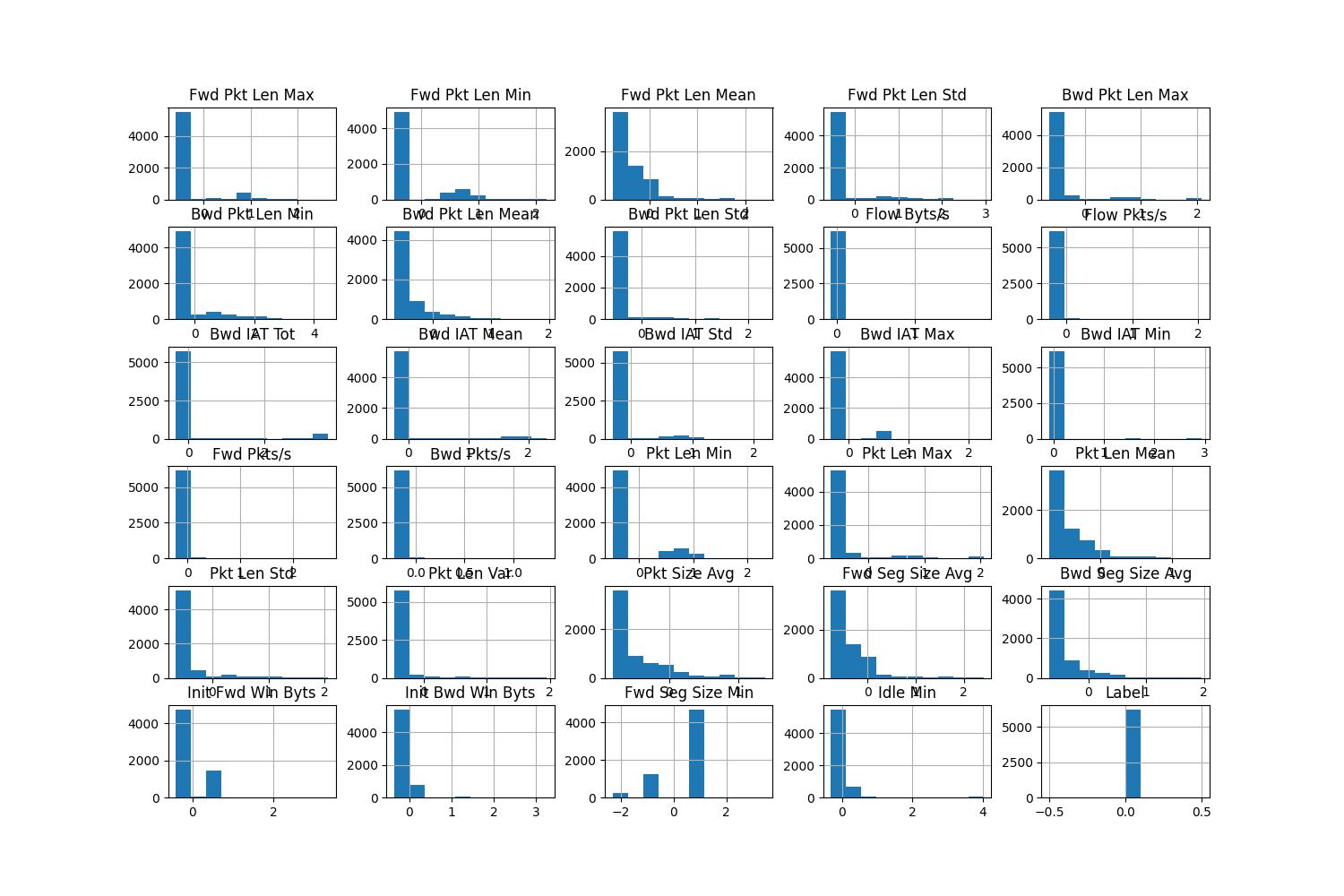} 
  \caption{Histogram of normal data features in Arrhythmia dataset.}
\end{figure}
\begin{figure}
  \centering
  \includegraphics[width=16cm,height=12cm]{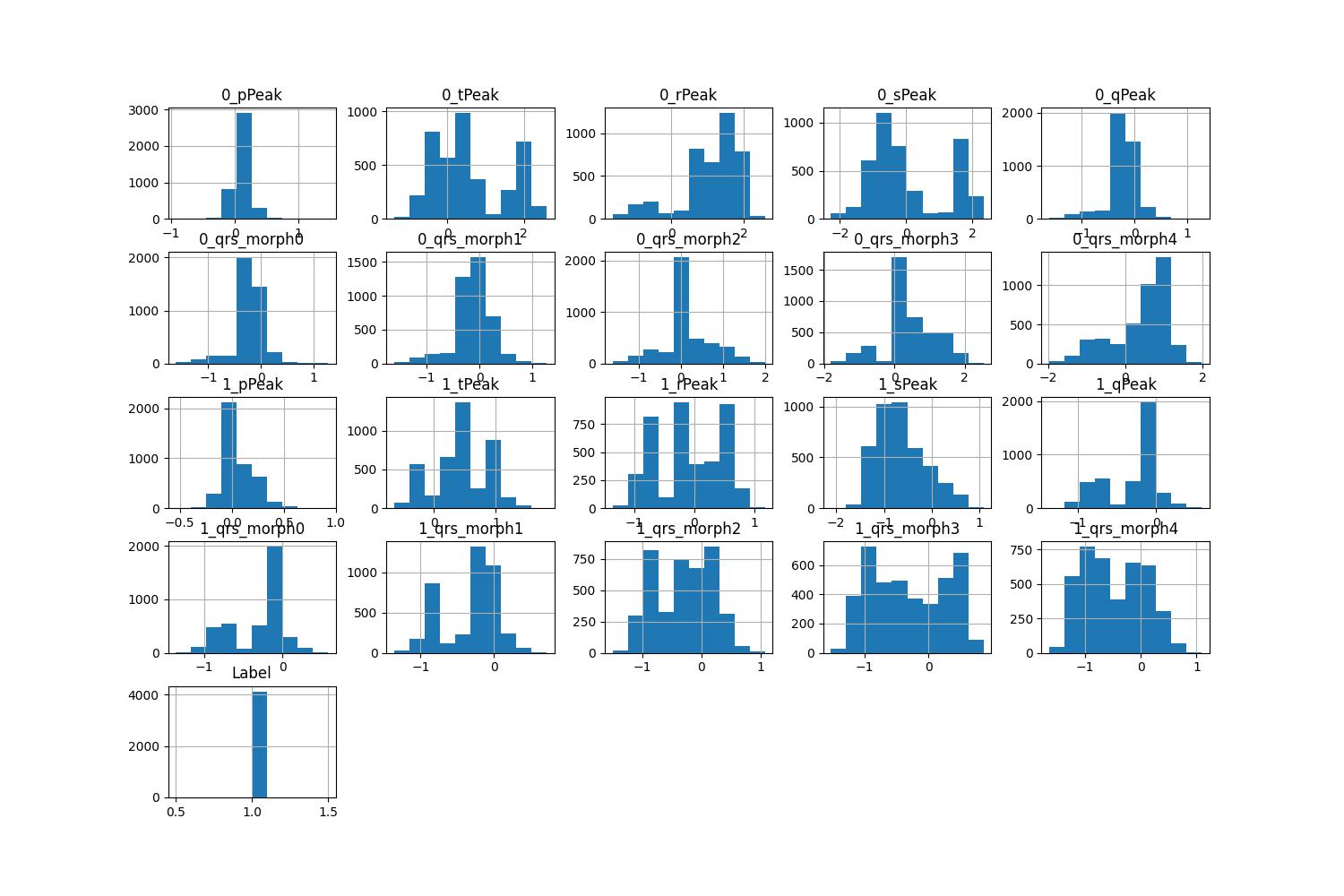}
  \caption{Histogram of far anomaly features(highly skewed) in Arrhythmia dataset.}
\end{figure}
\begin{figure}
  \includegraphics[width=16cm,height=12cm]{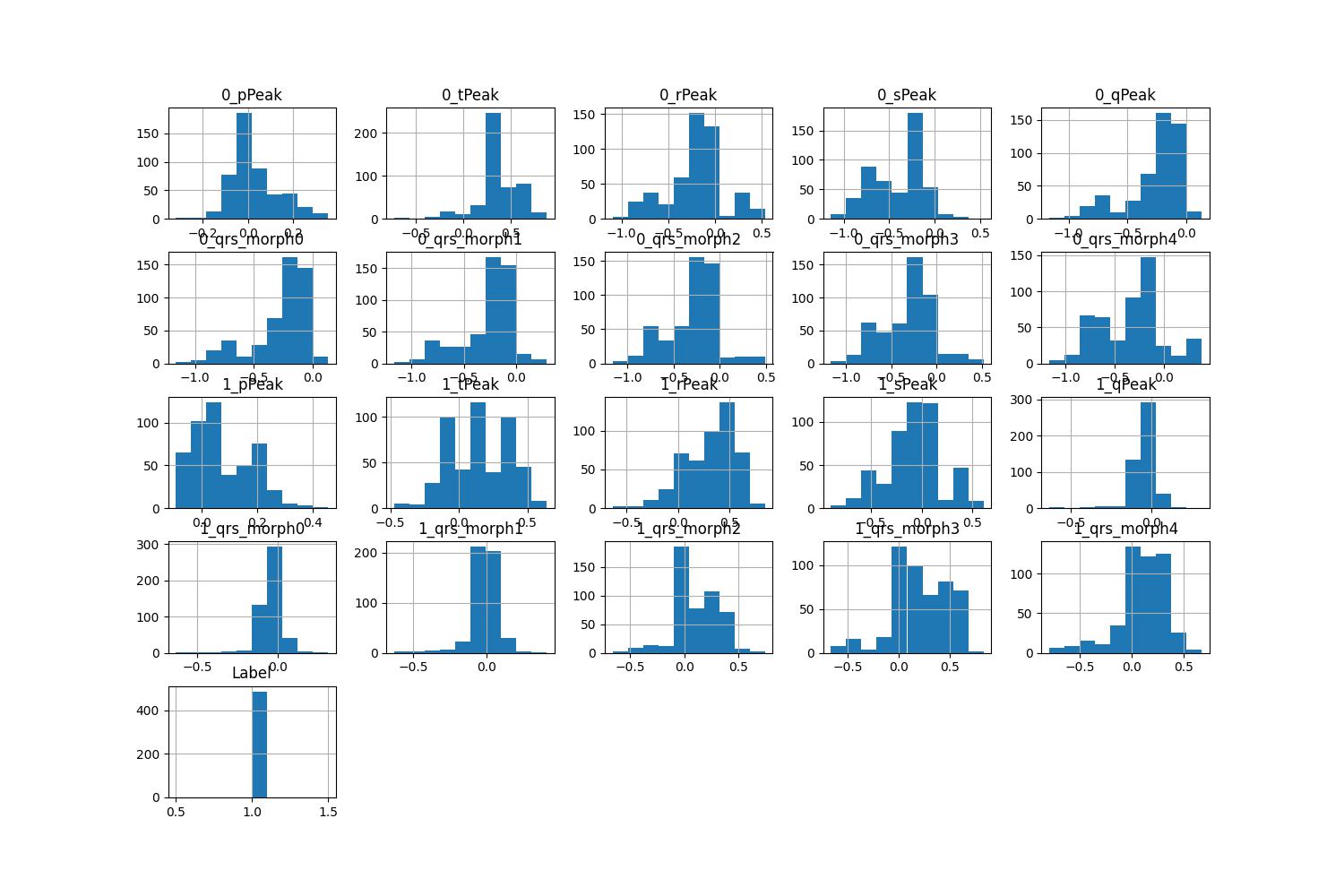}
  \caption{Histogram of near anomaly features(less skewed) in Arrhythmia dataset.}
\end{figure}
\begin{table}
\centering
\setlength{\tabcolsep}{1.5pt}
       \begin{tabular}{c|c|c}
        Features  & Skewness & Kurtosis 
        \\ \hline
        Fwd Pkt Len Mean &6.255196 &92.777752
        \\  \hline
        Flow Byts/s  &20.927526 &503.025265
        \\ \hline
        Bwd IAT Min &10.222297 &133.542522 
        \\ \hline
        Pkt Len Min &9.092836 &127.003666 
        \\ \hline
         Fwd Seg Size Avg &6.255196 &92.777752
        \\ \hline
         Bwd IAT Mean &15.838105 &133.542522
        \\ \hline
         Fwd Pkt Len Min &9.047784 &123.113359 
         \\
        \hline
        \end{tabular}
        \label{tab:results}
        \normalsize
\caption{The most skewed features in CSE-CIC-IDS dataset}
\end{table}
\begin{figure*}[htp!]%
  \subfloat[\centering]{{\includegraphics[height=2cm,width=3cm]{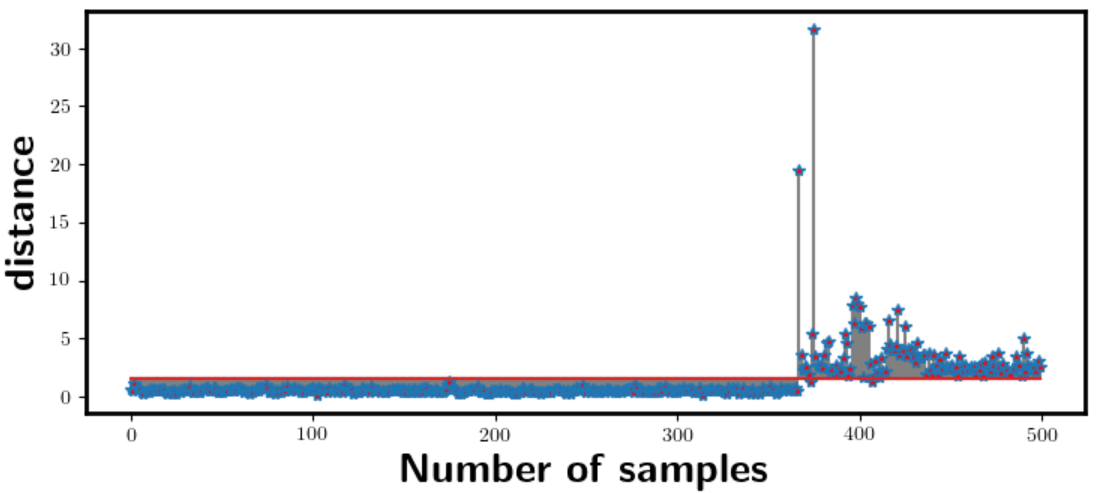}}}%
    \subfloat[\centering ]{{\includegraphics[height=2cm,width=3cm]{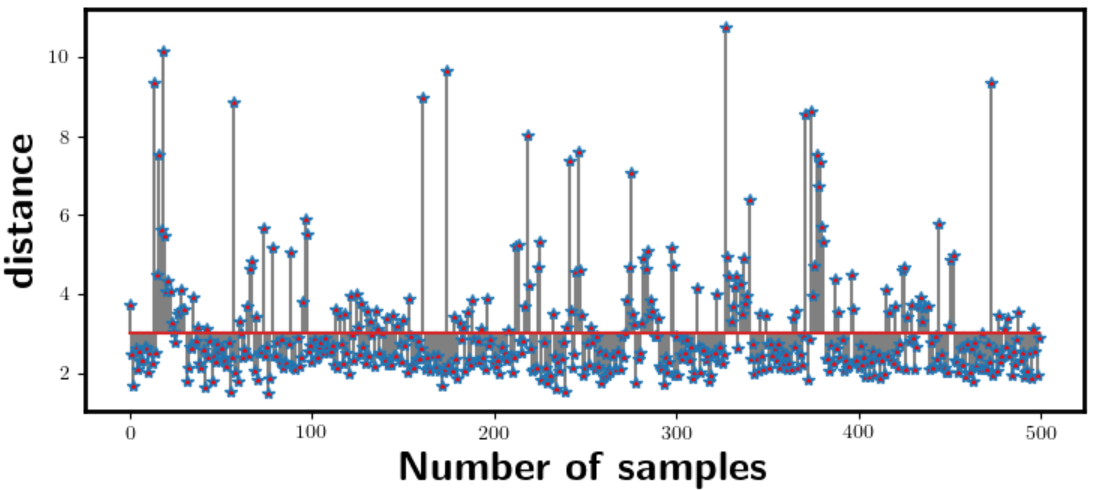}}}%
  \subfloat[\centering ]{{\includegraphics[height=2cm,width=3cm]{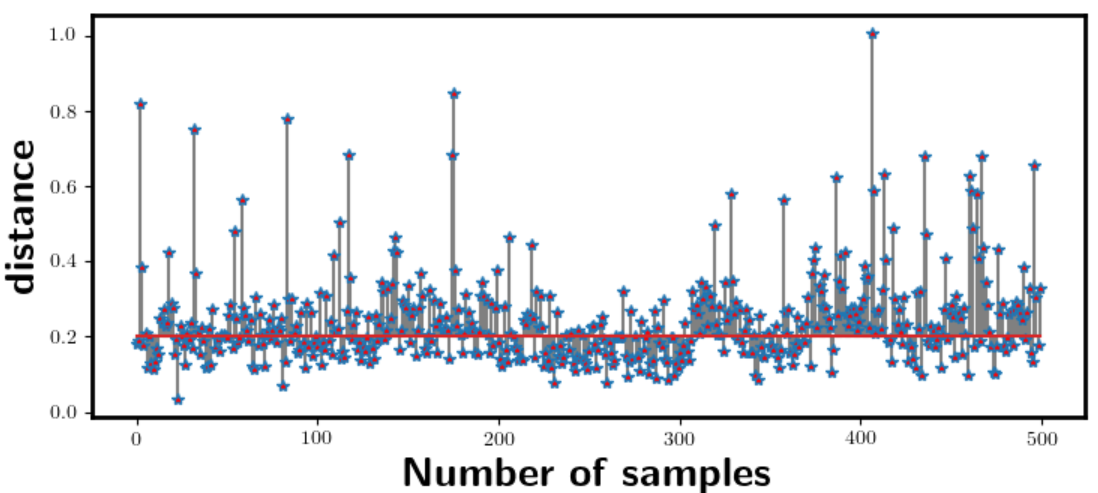}}}%
    \subfloat[\centering ]{{\includegraphics[height=2cm,width=3cm]{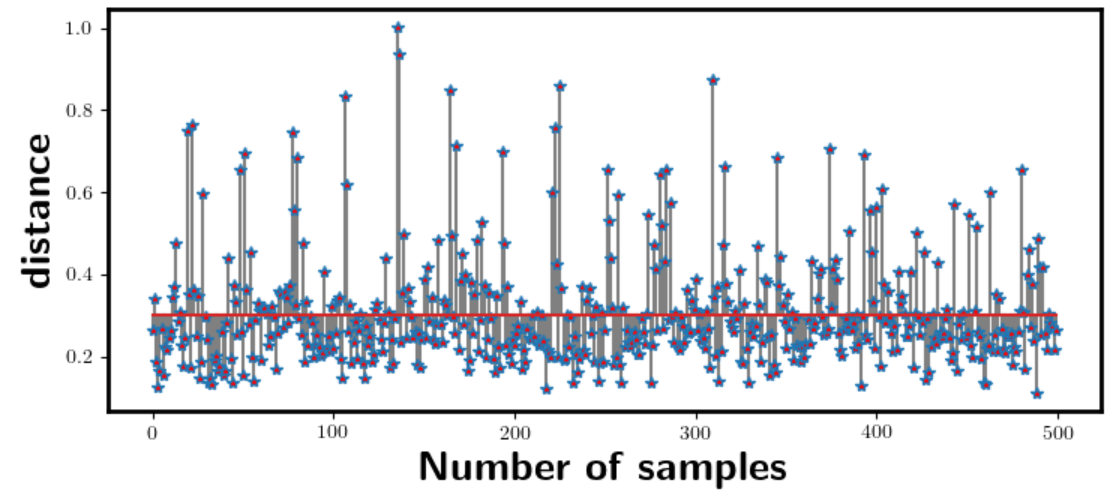}}}%
    \quad
    \quad
   \subfloat[\centering ]{{\includegraphics[height=2cm,width=3cm]{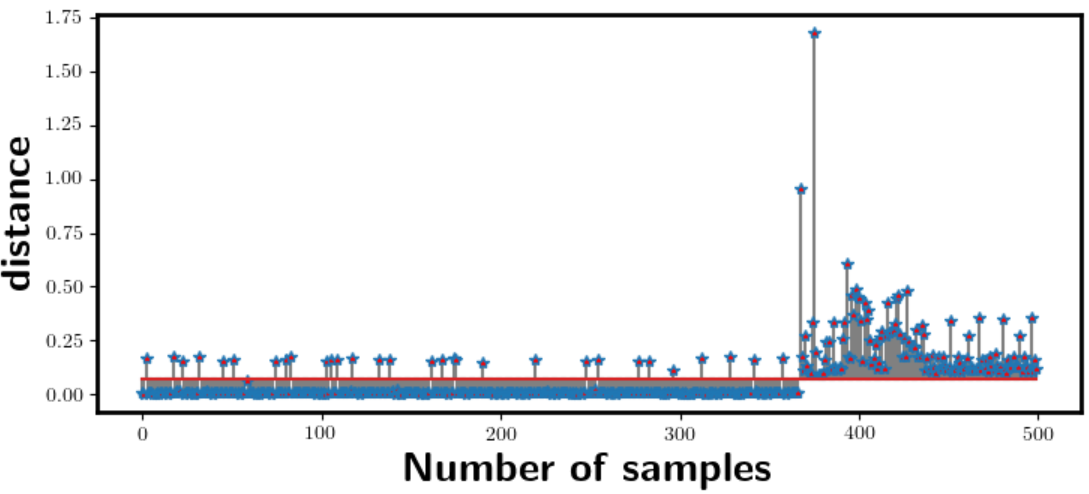}}}%
   \subfloat[\centering ]{{\includegraphics[height=2cm,width=3cm]{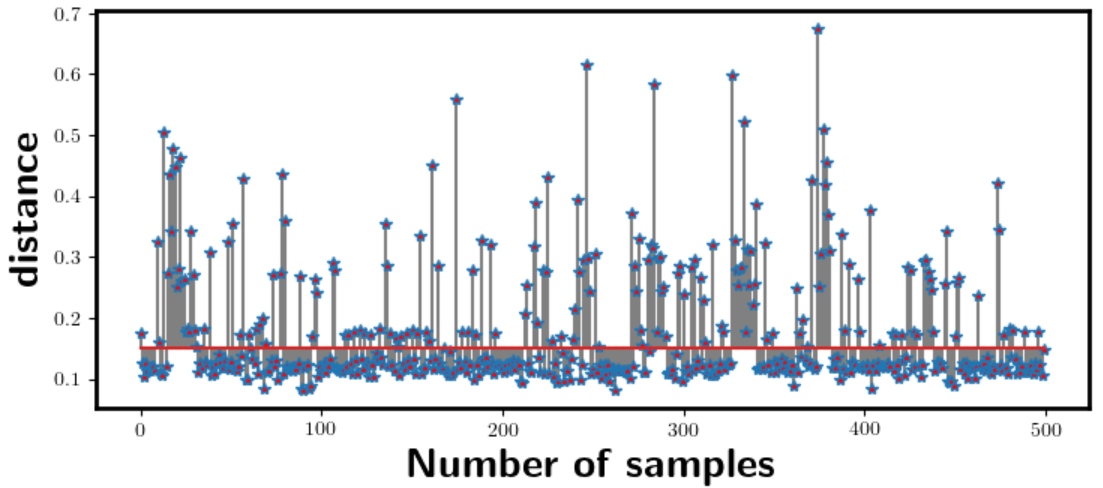}}}%
  \subfloat[\centering ]{{\includegraphics[height=2cm,width=3cm]{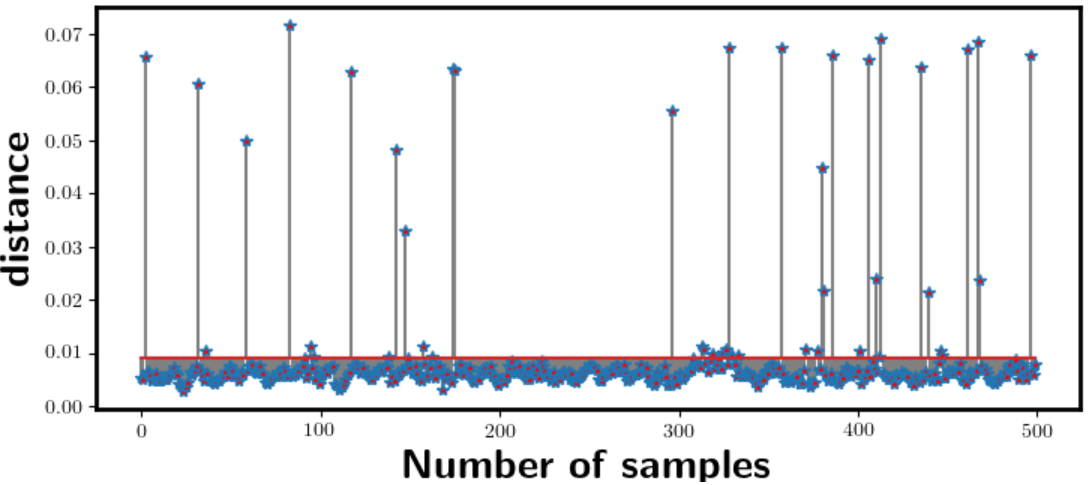}}}%
   \subfloat[\centering ]{{\includegraphics[height=2cm,width=3cm]{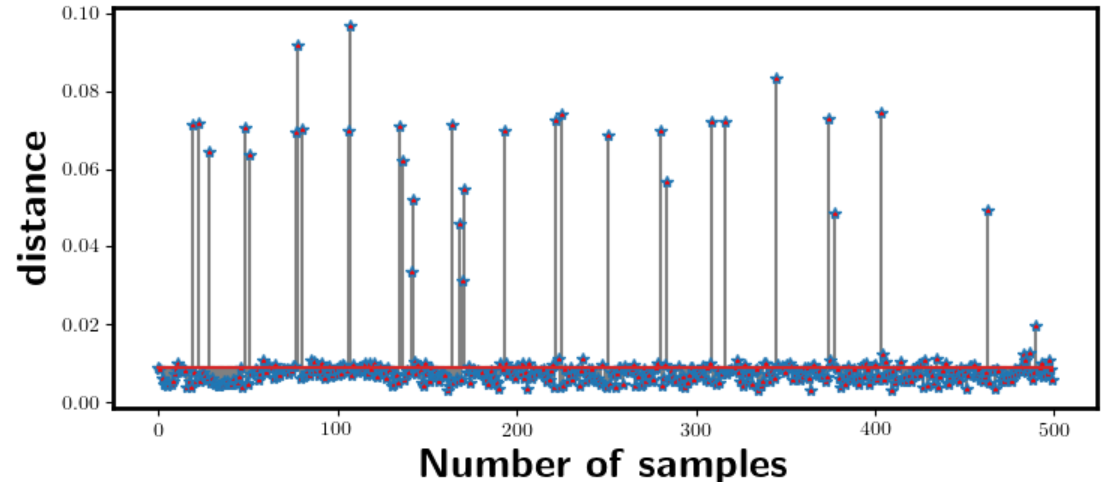}}}%
   \quad
   \quad
  \subfloat[\centering ]{{\includegraphics[height=2cm,width=3cm]{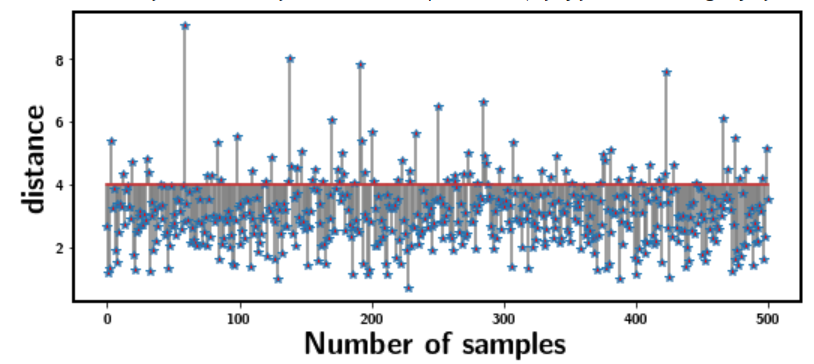}}}%
    \subfloat[\centering ]{{\includegraphics[height=2cm,width=3cm]{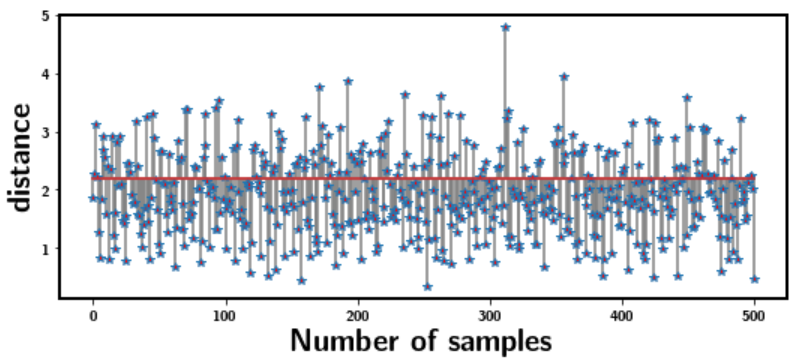}}}%
  \subfloat[\centering ]{{\includegraphics[height=2cm,width=3cm]{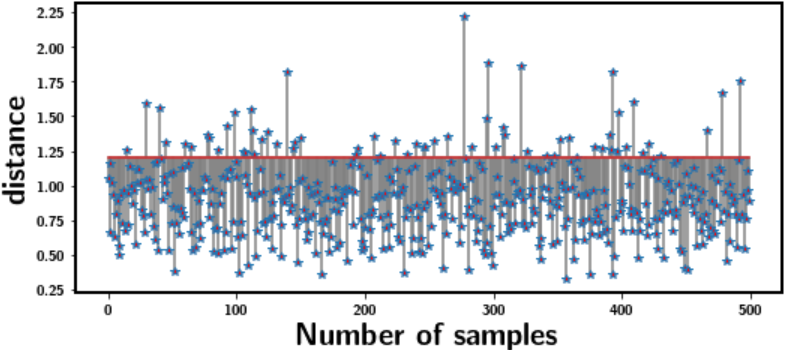}}}%
    \subfloat[\centering ]{{\includegraphics[height=2cm,width=3cm]{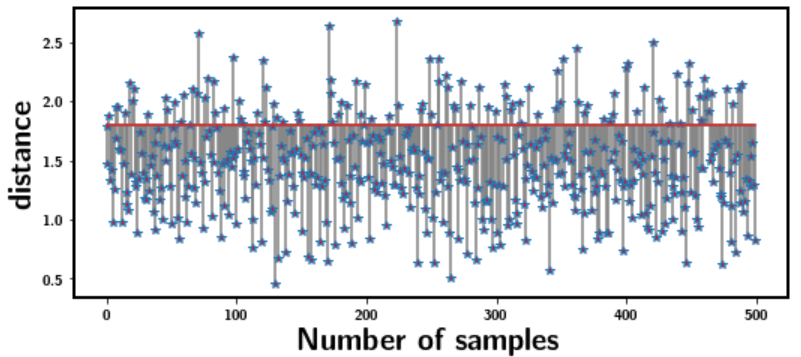}}}%
   \quad
   \quad
   \subfloat[\centering ]{{\includegraphics[height=2cm,width=3cm]{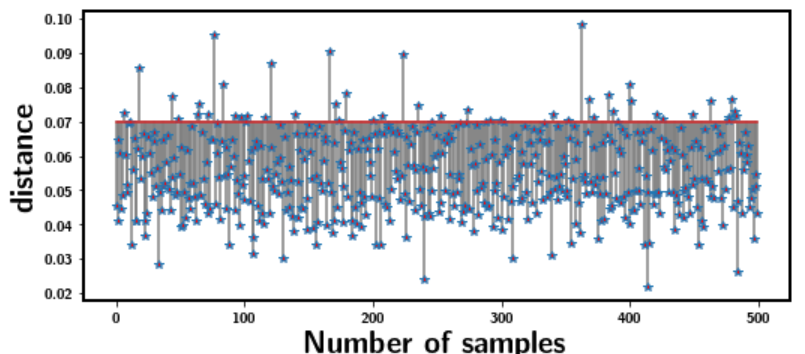}}}%
   \subfloat[\centering ]{{\includegraphics[height=2cm,width=3cm]{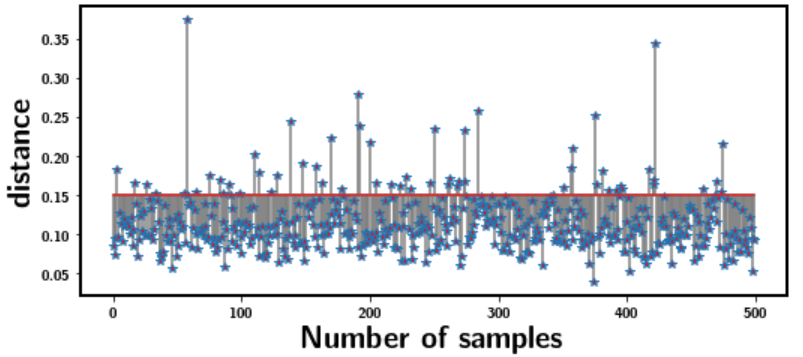}}}%
  \subfloat[\centering ]
  {{\includegraphics[height=2cm,width=3cm]{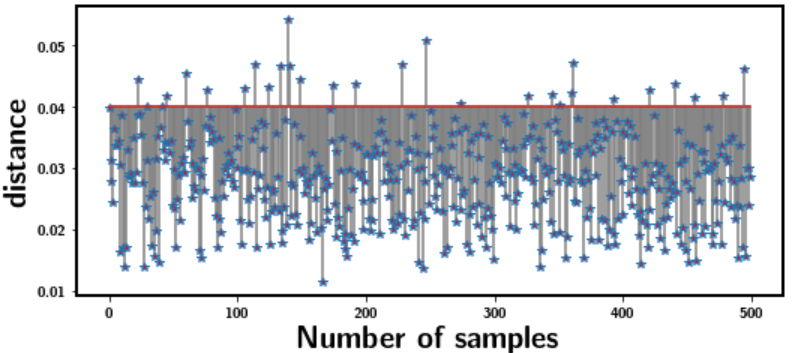}}}%
   \subfloat[\centering ]{{\includegraphics[height=2cm,width=3cm]{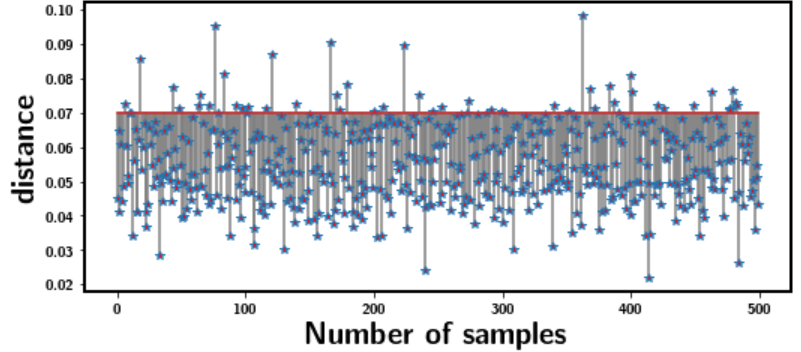}}}%
    \caption{ Fig (a-h) CSE-CIS-IDS data. 
 (a),(b) Detecting Far anomalies with DKAE;(c)(d)Detecting Near anomalies with DKAE; (e),(f) Detecting Far anomalies with DLSCA-AE; (g),(h) Detecting Near anomalies with  DLSCA-AE; Fig (i) to (p) shows the detection of near and far anomalies on the NSL-KDD dataset. The threshold is chosen based on the best value precision, recall, and accuracy of the test data.}
\label{Anomaly_detect_recons_space}
\end{figure*}

\end{document}